\documentclass{svproc}
\usepackage{graphicx}
\usepackage{url}

\usepackage{chngpage} 
\usepackage{float} 
\usepackage{makecell} 
\usepackage{booktabs} 
\usepackage{multirow} 
\usepackage{booktabs}  
\usepackage{array}
\usepackage{ragged2e}
\usepackage{lipsum}

\begin{document}
%
\title{Facial emotion recognition systems in smart classroom: A survey}
\titlerunning{FER technology and Smart classroom} 
\authorrunning{Amimi Rajae et al.}
\author{Amimi Rajae\inst{1}, Radgui Amina\inst{2}, and
Ibn el haj el hassane\inst{3}}

\tocauthor{Amimi Rajae, Radgui Amina, Ibn el haj el hassane}
\institute{National Institute of Posts and Telecommunications,Rabat, Morocco\\
\email{amimi.rajae@inpt.ac.ma}\\
\and
National Institute of Posts and Telecommunications, Rabat, Morocco\\\email{radgui@inpt.ac.ma}
\and
National Institute of Posts and Telecommunications,Rabat, Morocco\\
\email{ibnelhaj@inpt.ac.ma}\\}

\maketitle              

\begin{abstract}
Technology has transformed traditional educational systems
around the globe; integrating digital learning tools into classrooms offers
students better opportunities to learn efficiently and allows the teacher to
transfer knowledge more easily. In recent years, there have been many improvements in smart classrooms. For instance, the integration of facial
emotion recognition systems (FER) has transformed the classroom into an emotionally aware area using the power of machine intelligence and IoT. This paper provides a consolidated survey of the state- of-the-art in the concept of smart classrooms and presents how the application of FER systems significantly takes this concept to the next level.
\end{abstract}
\section{Introduction}
The concept of a modern classroom has long ago attracted the interest of many researchers. It dates back to the early 16th century when the Pilgrims Fathers established the first public school in 1635. Since the 1980s, with the development of information technology such as networking, multimedia,  and computer science, the classrooms of various schools have become more and more information-based at different levels.

Generally, a classroom is defined as an educational space where a teacher transfers knowledge to a group of students; this learning environment is one of the basic elements that influence the quality of education. Therefore, researchers suggest smart classrooms as an innovative approach that gives rise to a new intelligent teaching methodology, which became popular since 2012. The literature presents two visions of this concept.  The first approach that has taken good advantage of the joint growth of the computer science and electronics industry concentrates on the feasibility of deploying various intelligent devices in replacement to traditional materials, such as replacing books with optical discs or pen drives, getting rid of shalk board in favor of interactive whiteboard. In his paper  "what is a smart classroom?", Yi Zhang \cite{Zhang2019} notes '' The smart classroom can be classified as a classroom with computers, projectors, multimedia devices (video and DVD), network access, loudspeakers, etc., and capable of adjusting lighting and controlling video streams;''. According to some authors, this concept of a "technology-rich classroom" has a significant limitation in that it concentrates solely on the design and equipment of the classroom environment, ignoring pedagogy and learning activities \cite{williamson2017decoding}.  However, propositions from other studies point to a different insight, \cite{Kim2018} and al. envision making classrooms an emotionally aware environment that emphasizes improving teaching methodologies, this second approach focuses on the pedagogical aspect rather than technology and software design. Kim and al propose integrating machine intelligence and IOT to create a classroom that can listen, see and analyze students' engagement \cite{Kim2018} . 

In 2013, Derick Leony and al confirm that '' Many benefits can be obtained if intelligent systems can bring teachers with knowledge about their learner’s emotions" \cite{leony2013provision}, emotions may be a fundamental factor that influences learning, as well as a driving force that encourages learning and engagement. As a result,https://www.overleaf.com/project/62ec3930f99c128355b000c7 researchers suggest a variety of approaches for assessing students' affect states, including textual, visual, vocal, and multimodal methodologies. According to \cite {yadegaridehkordi2019affective}, the most widely utilized measurement channels are textual and visual. In comparison to the visual channel, textual (based on questionnaires and text analysis) is less innovative, and Facial Emotion Recognition (FER) systems are classified at the top of the visual channel.
In another state of the art, numerous researchers propose performant FER approaches with high accuracy up to 80\% \cite{Kas2021}; this encourages their integration as an efficient method to analyze student affect states.

Even though there has been an increasing amount of attention paid to technologies used in smart education, there is no literature that tackles the many elements of employing students' facial expression recognition systems in smart classrooms. Our article helps understand the concept of future classrooms and their technologies, particularly students' FER. 

We organize this paper as follows: In Section 2, we define smart classrooms and we present some of their technologies. In Section 3, we investigate the integration of FER systems to intelligent classrooms; enumerating FER databases and approaches. In Section 4,  we elaborate our insight and future works, then in Section 5, we present a brief conclusion.
\vspace{-3mm}
\section{Smart classroom: definition and technologies}
\subsection{Definition of smart classroom}
Smart learning is technology-enhanced learning, it facilitates interaction between students and their instructor and provides learners access to digital resources; it also provides guidance, tips, helpful tools, and recommendations to teach and learn efficiently. This innovative learning system comes with several new ways of learning classified into two categories: \\
\textit{Learning via technology:} for instance, taking interactive courses via massively open online courses (MOOC), educational games, or intelligent tutoring systems (ITS).\\
\textit{Learning with a teacher:} consists of learning in a technology-enhanced physical classroom, which is generally the concept of Smart classrooms (SC) .

\begin{figure}[h]
\begin{center}
\includegraphics [width=20pc ,height= 15 pc]{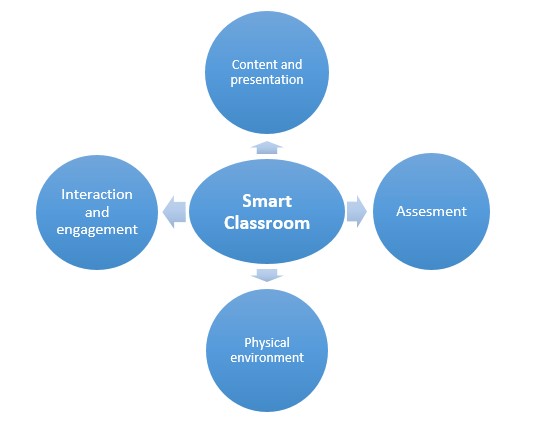}
\end{center}
\caption{Taxonomy of a typical smart classroom}
\label{fig: 1}
\vspace{-5mm}
\end{figure}

There are many definitions of smart classroom; indeed, it is hard to agree on a single definition that is accepted by the scientific community overall.
Authors like \cite{Zhang2019}, introduce intelligent classroom as learners-centered environment that supports students, adapts to their learning abilities, and helps teachers transfer their knowledge interactively and easily. Meanwhile, \cite{kwet2020smart} \cite{Saini2020} and al define a smart classroom as a physical environment containing digital tools, interactive devices, and various technologies to facilitate the activity of teaching and enhance learning. \cite{Kim2018}  and al comes up with a definition that goes beyond considering just the possibility of deploying intelligent materials into a physical place, they envision a smart classroom as an emotionally aware environment using real-time sensing and the power of machine intelligence, and in this context, they suggest a new system with advanced technologies and provide directions to deploy it.\\
In literature, authors present the concept of the future classroom in their own distinctive way, but the goal remains the same: to enhance education through technology.
A pertinent question to ask here, is, what elements make up a typical intelligent classroom? 
In response to this question, \cite{Saini2020} and al. propose a taxonomy of a typical smart classroom, as shown in Figure \ref{fig: 1}.

A typical smart classroom provides tools such as desktop computers, digital cameras, recording and casting equipments, interactive whiteboards, etc; \cite{Miraoui2018} for effective presentation, better assessment, constructive interaction, and comfortable physical environment. In order to present this taxonomy, four components are to consider: 
\\
\textbf{\textit{Smart content and presentation}:} technology assists the instructor in preparing the content of his courses and presenting it easily and interactively\cite{Miraoui2018}.
\\
\textbf{\textit{Smart assessment}:}  includes automated evaluation of students learning capacity; also, it consists of managing students' attendance and recuperating their feedback to enhance lectures' quality as presented by \cite{Tang2015} \cite{Ashwin2019}.
\\
\textbf{\textit{Smart physical environment}:} smart classrooms offer a healthy climate by controlling factors like air, temperature, humidity, etc; using sensors and actuators \cite{Pacheco2018} \cite{revathi2020iot} \cite{FarithaBanu2020} . 
\\
\textbf{\textit{Smart interaction and engagement}:} it focuses on analyzing the level of students engagement and consists of providing tools to enhance interaction \cite{Tang2015} \cite{Kapoor2007} \cite{Kapoor2005}.

\subsection{Smart classroom technologies}
Education technology (edtech) is a multi-billion-dollar business that is rising every year. Many nations in the Organisation for Economic Co-operation and Development (OECD) spend more than 10\% of their budgets on education \cite{khosravi2021learning}.
Furthermore, as more nations raise their education investment, public spending on Edtech is expected to rise. Low- and middle-income nations, for example, expect to boost education expenditure from US\$ 1.2 trillion to US\$3 trillion per year \cite{declaration2016sdg4}. According to the Incheon Declaration, nations must devote at least 4\% to 6\% of their GDP to education, or at least 15\% to 20\% of public spending to education.
Aside from the predicted expansion in the educational sector, the market for smart classroom technologies is rapidly growing and is strongly linked to advances in computer science, robotics, and machine intelligence.
As indicated in Table \ref{tab: 1}, every technology has advantages and disadvantages; the most prevalent constraints of most of these technologies are cost in the first position, followed by technical knowledge concerns ( generally, teachers and students don't have the required technical knowledge to use those technologies correctly) in the second place. Below are some of the main SC technologies: \\
\textbf{Interactive Whiteboard} (IWB): is an intelligent tool that allows users to manipulate their presentations and project them on a board's surface using a special pen or simply their hand \cite{Lant2016}. IWB can be used to digitalize operations and tasks or merely as a presentational device \cite{Glover2005}. The use of this device has revolutionized the nature of educational activities; it has the power to reduce the complexity of teaching and offers the instructor more flexibility during presentations. \\
\textbf{RFID Attendance Management System}: RFID stands for Radio Frequency Identification; it is a wireless technology used to track an object then memorize and recuperate data using radio tags \cite{Hasanein2018}. An RFID attendance system automatically marks students' presence in the classroom by validating their ID cards on the reader. It is a bright, innovative solution to replace classic attendance registers \cite{Kassim2012}, and help teachers gain wasted time verifying students' presence daily.\\
\textbf{Educational cobot}: is a new innovative tool used as a co-worker robot to help with teaching tasks. It contains several sensors, cameras, microphones, and motors so it can listen, see, communicate with students and assist the teacher \cite{Alimisis2017}. Embedded into classrooms, cobots represent the school of the future and have significantly changed the traditional ways of teaching \cite{Timms2016}.\\ 
\textbf{Sensors and actuators}: consists of installing sensors to collect data from their environment, then sending this data to the cloud to be analyzed, and might decide to act using actuators; for example, a temperature sensor detects and measures hotness and coolness, then sends the information to another device to adjust the temperature. This technology provides an adequate healthy climate by controlling air, temperature, humidity, etc \cite{revathi2020iot}.\\
\textbf{Augmented Reality} (AR): is a system that combines real and virtual worlds, it is a real-world interactive experience in which computer-generated perceptual information enhances real objects \cite{akccayir2016augmented}. It was first used in training pilots applications in the 1990s \cite{thomas1992augmented}, and over the years, it has demonstrated a high efficiency when adopted in educational settings; it has been used to enhance many disciplines particularly when students learn subjects with complex, abstract concepts or simply things they find difficult to visualize such like mathematics, geography, anatomy, etc \cite{Chen2016}.\\
\textbf{A Classroom Response System} (CRS): is used to collect answers from all students and send them electronically to be analyzed by the teacher, who can graphically display a summary of the gathered data \cite{2019}. CRS is an efficient method to increase classroom interaction \cite{Fies2006}.According to Martyn (2007) \cite{martyn2007clickers}, “One of the best aspects about an CRS is that it encourages students to contribute without fear of being publicly humiliated or of more outspoken students dominating the discussion.”\\
\textbf{Commercial off-the-shelf} (COTS) eye tracker : is a gaze-based model that is used to monitor students' attention and detect their mind wandering in the classroom. It helps the instructor to better understand their interests and evaluate their degree of awareness \cite{Hutt2017} \cite{Hutt2019}.\\
\textbf{Wearable badges}: tracks the wearer's position, detects when other badge wearers are in range, and can predict emotion from the wearer's emotional voice tone. It helps the teacher manage the classroom; he can use those badges to detect if a student leaves without permission or when a group of learners make a loud noise that disturbs the rest of their classmates. MIT has developed a wearable badge by Sandy Pentland's team \cite{Timms2016} \cite{khosravi2021learning}.\\
\textbf{Learning Management System} (LMS): is a web-based integrated software; used to create, deliver and track courses and outcomes. It aids educators to develop courses, post announcements, communicate easily and interactively, grade assignments, and assess their students; besides, it allows students to submit their work, participate in discussions, and take quizzes \cite{Courts2012} \cite{Conde2012} \cite{alhazmi2012lms}. 
\\
\textbf{Student facial emotions recognition} (FER): is a technology that analyzes expressions using a person's images; it is part of the affective computing field. Authors uses this technology to predict real-time student feedback during lectures \cite{Kas2021}, which helps the instructor improve the quality of his presentations.

\vspace{-5mm} 

\begin{center}
\begin{table}[ht]
\caption{ Merits and limitations of smart classroom technologies}
\begin{adjustwidth}{-0.5in}{1in}

\scalebox{0.7}{
\newcolumntype{c}{>{\raggedcolumns\arraybackslash}p{8cm}}
\newcolumntype{r}{>{\raggedcolumns\arraybackslash}p{9cm}}
\newcolumntype{r}{>{\raggedcolumns\arraybackslash}p{9cm}}

\begin{tabular}[ht]{m{0.2\textwidth}|m{0.7\textwidth}|m{0.7\textwidth}}


\hline

\thead{Technology}  & \thead{Merits}  & \thead{Limitations } \\ 

\hline
IWB  & \thead{The touchscreen made its use simpler and more effective.  \\
      It is equipped with smart tools such as a pointer, screen capture...\\
      It provides access to the web.} & \thead{ Expensive: many schools are not able to afford it.  \\
          Teacher training: The school should spend time and  money teaching their instructors   how to  use the equipment correctly. } \\

\hline
RFID  & \thead{Quick and Rapid: it identifies students in seconds.\\
Accuracy : it provides more accurate identification. } & \thead{ Expensive : In case of a large strength of students,\\ purchasing tags for everyone is costly.
\\Not secure : the system  is prone to manipulation.
 } \\

\hline
COBOT & \thead{Wide Knowledge: it saves a large amount of information.}  &  \thead{Technical Disruptions: it can break down at any time.\\
Human-machine interaction: it may have trouble in trying to interact with the students.\\
Expensive. } \\

\hline
Sensors and actuators & \thead{Comfort: it provides a healthy and comfortable environment.} & \thead{Technical support : it needs  IT professionals to help set it up and maintain it } \\

\hline
AR & \thead{It provides outstanding visualizations.\\
It Increases Students' Engagement.} & \thead{It may presents functionality Issues.\\
It is expensive.} \\

\hline
CRS & \thead{Rapid assessment: It provides outcomes of formative assessment. \\It provides immediate feedback for student.\\
Time saving: for instance, Fast grading.} & \thead{Expensive : it costs an average of \$75 / device\\
It presents technical problems.\\
Ineffective for opinion questions.} \\

\hline
COTS & \thead{It records real time eye movements and fixations, which can report student reactivity.} & \thead{It is not able to track all eyes: eye-tracking camera is \\impacted for example by lenses or glasses.\\
It costs money, time and labor resources. \\
Efficiency:  visual attention is not sufficient to interpret \\ students’ engagement.
} \\

\hline
\thead{Wearable\\ badges} & \thead{It increases the level of security and privacy in classrooms.} & \thead{Privacy concern: pupils may have some misapprehension\\ about their privacy when it comes to wearable devices.} \\

\hline
LMS & \thead{It saves wasted time on menial tasks like grading papers.\\
It gives students access to learning material in one place from any device.} & \thead{It requires IT and programming knowledge.} \\

\hline

\end{tabular}}

\end{adjustwidth}
\label{tab: 1}
\end{table}
\end{center}


\section{ The application of FER system to smart classroom}

\subsection{FER approaches used for smart classroom }
Analyzing student’s affective states during a lecture is a pertinent task in smart classroom \cite{gligoric2012smart}. It is crucial to determine student engagement during a lecture in order to measure the effectiveness of teaching pedagogy and enhance the interaction with the instructor. Therefore, in order to get student feedback and achieve his satisfaction, researchers suggest different methods such as: body gesture recognition detected by using Electroencephalography (EEG) signals \cite{Kumar2016}, body posture using either cameras or a sensing chair\cite{Kapoor2005}, hand gestures \cite{Ashwin2020a}, heart rate \cite{Monkaresi2017}, and so forth. Meanwhile, FER systems appear in literature like the most used solution to recognize student’s affective states in smart classroom.

\begin{figure}
\begin{center}
\includegraphics [width=28pc ,height= 15pc]{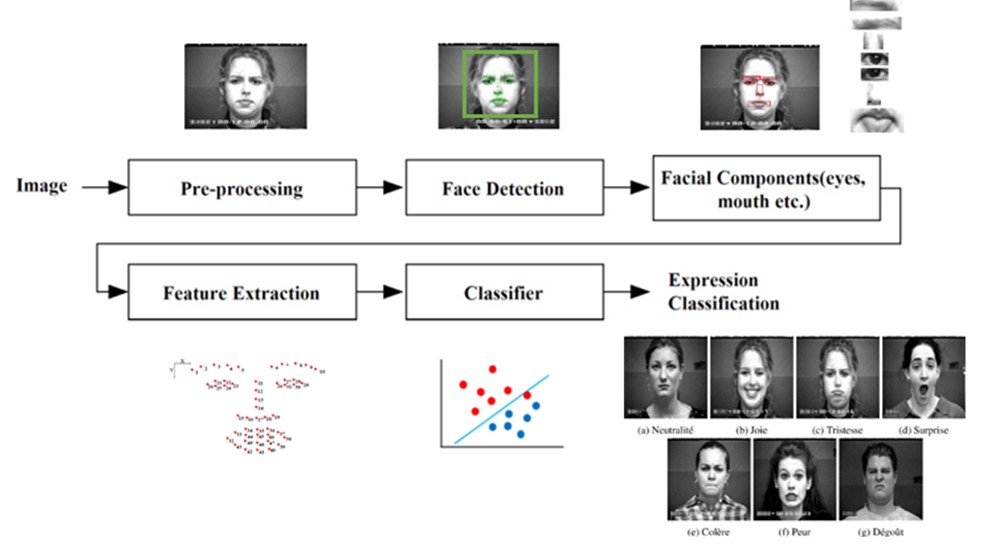}
\end{center}
\caption{Facial expressions recognition system Framework }
\label{fig: 2}
\end{figure}

Recently, FER systems are used in several domains like robotics, security, psychology, etc; thus, over time many researchers suggest new performant FER approaches. The framework of FER systems is structured as shown in Figure \ref{fig: 2}; the input of this system is the images from a FER database; those images are pre-processed in a way to match the input size of the network, then, adequate algorithms are used to detect area of the face properly and extract the main features that help the network learn from training data; the final step is to classify results according to the database labels \cite{Kas2021}. Thousands of articles have been written on this subject, but only a few of them have applied FER systems to smart classrooms. 
\vspace{-4mm} 

In Table \ref{tab: template}, we cite FER approaches used in smart classrooms; we classify those approaches into three categories: \\
\textbf{Handcrafted approaches}: are traditional methods of machine learning that consist of manually extracting features. Some examples include edge detection, sharpening, corner detection, histograms, etc. LBP pattern, for instance,  is a type of image descriptor used to extract a texture of an image. Then, for classification, an adequate classifier is used; it is also a traditional machine learning algorithm such as: Support Vector Machine (SVM), K-nearest neighbor (KNN), decision tree, etc.\\
\textbf{Deep learning approaches}:  are new methods based on neural networks for both feature extraction and classification. Learned features are extracted automatically using deep learning algorithms; Convolutional Neural Network (CNN) is the most commonly used for analyzing visual imagery; it can choose the best features to classify the images.  \\
\textbf{Hybrid approaches}: combine both algorithms of machine learning and deep learning. They utilize traditional machine-learning algorithms to extract features and neural networks to classify images;
\vspace{-5mm} 


\begin{table}[H]
\caption{FER approaches in application to smart classroom}
\begin{adjustwidth}{0.1in}{1in}
\begin{tabular}{c|c|c|c|c|c}
\hline
Approach & Works & Year &\thead{Feature \\ extraction  } & Classification & Accuracy\\
\hline
\multirow{2}{*}
        {Handcrafted \vspace{-4mm} } & \cite{Whitehill2014} & 2014 &	Gabor features & \thead{Support vector \\ machine (SVM) } &    72\% \\
\cline{2-6}
& \cite{Tang2015}& 2015 & ULGBPHS & \thead{K-nearest neighbor  \\ classifier (KNN)} & 79\% \\
\hline
Hybrid & \cite{Kaur2018} & 2019 & LBP-TOP	& \thead{Deep Neural Network \\(DNN)}  & 85\% \\
\hline
\multirow{3}{*}
        {Deep Learning \vspace{-20mm} } &	\cite{Lasri2019} & 2019	& \multicolumn{2}{c|}{\thead{Convolutional Neural Network \\ (CNN)	}}  & 70\% \\
\cline{2-6}
& \cite{AshwinT.2019} & 2020 &	\multicolumn{2}{c|}{ \thead{CNN-1: analyze single face \\expression in single image.\\
CNN- 2 analyze multiple faces\\ in single image }}
&  \thead{86\% for CNN1 \\ 70\% for CNN2 } \\
\cline{2-6}
& \cite{Ashwin2020a} & 2020	&
\multicolumn{2}{c|}{\thead{CNN based\\ on GoogleNet architecture\\ with 3 types of Databases}} & \thead{88 \% \\79 \% \\61\%} \\
\hline
\end{tabular}
\end{adjustwidth}
\vspace{-7mm} 
\label{tab: template}
\end{table}

\paragraph{Notes and Comments.} In the beginning, approaches (before 2019) employed classical machine learning methods like Gabor filters, LPB, KNN, and SVM to extract features and classify emotions; then gradually, since 2019, authors started integrating neural networks. As shown in Table \ref{tab: template}, \cite{Kaur2018} and al utilize a hybrid approach that consists of using LBP-TOP as a descriptor then deep neural network (DNN) for classification which gives a better result (85\%) in comparison to both the methods proposed by  \cite{Whitehill2014}\cite{Tang2015} accuracies (respectively 72\%, 79\% ) and the new deep method suggested by \cite{Lasri2019}; additionally, Ashwin and al. suggest methods with different accuracies improved ( from 61\% to 88\%) by changing the training data \cite{AshwinT.2019} \cite{Ashwin2020a}; which point on the importance of choosing an adequate database.

\subsection{FER student’s databases}
Over the past years, there was a lack of student facial expressions databases \cite{gligoric2012smart}, and most of the authors use general FER databases like FER2013 and CK+ to train their models \cite{Ashwin2020}\cite{Ashwin2020a}.
It is essential to have an adequate database to increase the models' accuracy and get better results [16], because the efficiency of FER models depends mainly on the quality of both databases and FER approaches [19]. 
Each created database has its characteristics, depending on the classes of expressions, ethnicity of participants, labeling methods, Size, method and angle of image acquisition. In Table \ref{tab: 2}, we have gathered students' expressions databases used in intelligent classrooms.
\clearpage


\begin{table}[ht]
\caption{ Student’s FER databases}
\begin{adjustwidth}{-0.4in}{1in}
\scalebox{0.8}{
\begin{tabular}{c|c|c|c|c|c|c}

\hline
Works & Database & \thead{Expressions\\ Classes} & \thead{Ethnicity \& \\ gender of \\ participants} & \thead{Method or angle \\ of acquisition} & Labeling & \thead{Database \\ content } \\

\hline
\cite{Whitehill2014} & Spontaneous & \thead{4 classes\\ Not engaged \\ Nominally engaged \\ Engaged in task \\Very engaged} & \thead{Asian-American\\  Caucasian-American  \\ African-American  \\  25 female} & \thead{Pictures are taken\\ from an IPad\\ camera posed\\ in 30 centimeters\\ in front of the\\ participant’s face\\ while playing\\ a cognitive game} & \thead{Labeled by\\ students from\\ different disciplines:\\ computer science,\\  cognitive and\\ psychological science} & \thead{ N/A\\ 34 participant} \\

\hline
\cite{Tang2015} & Spontaneous & \thead{5 classes: \\Joviality\\ Surprise\\ Concentration\\ Confusion\\ Fatigue } & Asian & \thead{ Acquired using a\\ full 1080p HD\\ camera configured\\ at the front of\\ the classroom while\\ the student watching\\ a 6 minutes video } & \thead{ Participants labeled\\ their own pictures } & \thead{ 200 images\\ 23 participants} \\

\hline
\cite{ZatarainCabada2017} & Spontaneous & \thead{4 Classes :\\ Frustration\\ Boredom\\ Engagement\\ Excitement} & N/A &  \thead{Computer webcam\\ takes a photograph\\ every 5 seconds} & \thead{Labeled using\\ a mobile\\ electroencephalo\\-graphy (EEG)\\ technology called\\ Emotiv Epoc} & 730 images \\

\hline
\cite{Bian2019} & Spontaneous & \thead{5 classes:\\ Confusion\\ Distraction\\ Enjoyment\\ Neutral\\ Fatigue} & \thead{Chinese\\ (29 male and\\ 53 female)} & \thead{Acquired using\\ Computers cameras} & \thead{Labeled by\\ the participants\\ and external coders} & \thead{1,274 video\\ 30,184 mages\\ 82 participants} \\

\hline
\cite{Ashwin2020a} & \thead{Posed and\\ spontaneous} & \thead{14 Classes:\\ 7 classes of\\ Ekman’s basic\\ emotions\\ 3 learning-centered\\ emotions\\ ( Frustration,\\ confusion\\ and boredom )\\ Neutral} & Indian & \thead{Frontal posed\\ expressions} & \thead{Labeled using\\ the semi-automatic\\ annotation process\\ and reviewed\\ manually to\\ correct wrong\\ annotations} & 4000 \\

\hline
\end{tabular}}
\end{adjustwidth}
\label{tab: 2}
\end{table}
\vspace{-8mm} 

\section{Insights and future work}
In this paper, we have surveyed the concept of smart classroom and its technologies, especially FER systems. In the future, we plan to consider the evolution of smart classrooms in this critical period of the pandemic.

\begin{figure}
\begin{center}
\includegraphics [width=25pc ,height= 16pc]{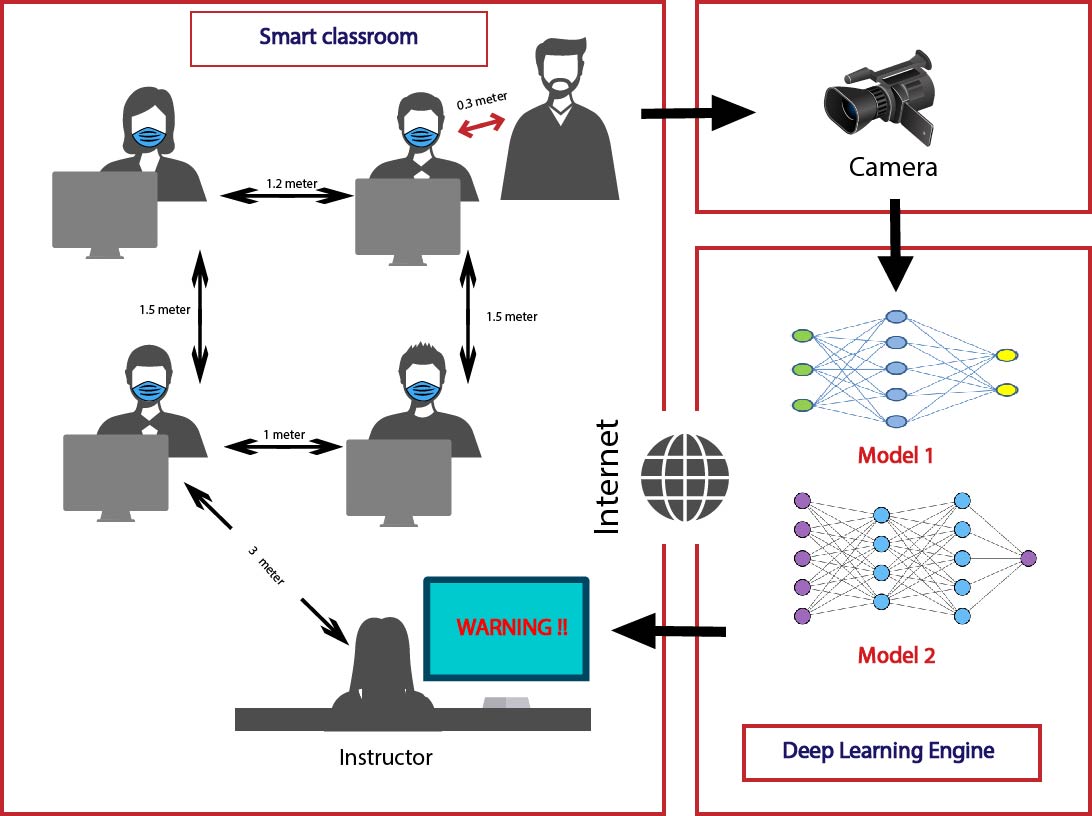}
\end{center}
\caption{Proposed model for intelligent classroom }
\label{fig: 3}

\end{figure}

Today, the pandemic of coronavirus is causing a global health crisis. During this time, countries around the world have imposed restrictions on social distancing, masking, and other aspects of public life. The lockdowns in response to the spread of the virus are significantly impacting educational systems. As a result, governments are striving to maintain continuity of learning and are proposing distance learning as a suitable interim solution, but unfortunately, not all students around the globe have access to digital learning resources. In our future work, we propose a model for an intelligent physical classroom that respects the restrictions of Covid-19. We base our model on two propositions; as shown in Figure \ref{fig: 3}:

\textbf{Proposition 1}: we propose a system that automatically detects whether the students in the classroom are wearing their masks or not.

\textbf{Proposition 2}: we detect the distances between students and compare them with the allowed distance. 

If the students do not comply with the restrictions, the system sends a warning to the teacher in real-time.

\vspace{-8mm} 
\paragraph{Notes and Comments.} The Proposed intelligent classroom system in Figure \ref{fig: 3} consisting of (Model 1, Model 2) artificial neural networks to detect students not wearing masks and to measure distances between learners.\\
Wearing face masks strongly confuses facial emotions recognition systems (FER). In our future work, we will study the limitations of these systems, used in today's smart classrooms, as well as the possibility of predicting emotions in a student face wearing a mask;  since it is possible, for example, to predict students' mind wandering and the level of their engagement based only on their gaze \cite{Hutt2017} \cite{Hutt2019}.

\section{Conclusion}
Smart classroom is not a new concept, but over the years it has known many changes through the integration of various technologies.
Researchers have transformed the classroom from a simple physical space gathering learners and their instructors to an emotionally aware environment that can interact with students and help them learn efficiently. Numerous technologies have revolutionized the evolution of digital classrooms, and FER systems are considered the most innovative.
 The authors have adapted FER 's systems for use in smart classrooms using approaches with high accuracies and personalized databases.\\
The evolution of these interactive classrooms can be considered to aid in teaching during the restrictions of the COVID 19 pandemic. It can also be adapted for teaching students with special needs or intellectual disabilities to facilitate their interaction with their teachers. 
\section{Acknowledgement} 
Authors would like to thank the National Center for Scientific and Technical
Research (CNRST) for supporting and funding this research.


\bibliography{biblio_refrence_rajae}

\end{document}